\documentclass{aircc}
\usepackage{mathpartir}
\usepackage{hyperref}
\usepackage{listings}
\usepackage{latexsym}

\usepackage[T1]{fontenc}
\usepackage[utf8]{inputenc}

\usepackage{microtype}
\usepackage{url}
\newcommand\R{\mathbb{R}}
\usepackage{graphicx}
\usepackage{multirow}
\usepackage{booktabs}
\usepackage{amsmath,amssymb}
\usepackage{array}
\newcolumntype{P}[1]{>{\centering\arraybackslash}p{#1}}
\usepackage{url}
\usepackage{xcolor}
\usepackage{caption,subcaption}
\usepackage{courier}
\usepackage{soul}
\usepackage{colortbl}
\usepackage{pgffor}
\usepackage{etoolbox}
\usepackage{longtable}
\definecolor{r}{RGB}{240,160,180}

\title{Evaluating and Improving\\ Context Attention Distribution on\\ Multi-Turn Response Generation using\\ Self-Contained Distractions}

\author{Yujie Xing and Jon Atle Gulla}
\affiliation{Norwegian University of Science and Technology \\ \email{\{yujie.xing, jon.atle.gulla\}@ntnu.no}}

\begin{document}
\maketitle
\begin{abstract}
Despite the rapid progress of open-domain generation-based conversational agents, most deployed systems treat dialogue contexts as single-turns, while systems dealing with multi-turn contexts are less studied. There is a lack of a reliable metric for evaluating multi-turn modelling, as well as an effective solution for improving it. In this paper, we focus on an essential component of multi-turn generation-based conversational agents: \textbf{context attention distribution}, i.e. how systems distribute their attention on dialogue's context. For evaluation of this component, We introduce a novel attention-mechanism-based metric: \textbf{DAS ratio}. To improve performance on this component, we propose an optimization strategy that employs self-contained distractions. Our experiments on the Ubuntu chatlogs dataset show that models with comparable perplexity can be distinguished by their ability on context attention distribution. Our proposed optimization strategy improves both non-hierarchical and hierarchical models on the proposed metric by about $10\%$ from baselines.
\end{abstract}


\section{Introduction}

In recent years, generation-based conversational agents have shown a lot of progress, while multi-turn generation-based conversational agents are still facing challenges. Most recent work ignores multiturn modelling by considering a multi-turn context as a 1-turn context \cite{dialogpt,knowledge2020}. Some works try to deal with multi-turn modelling using modified attention mechanisms, hierarchical structures, utterance tokens, etc. \cite{iulian1,iulian2,dialogflow}. The main difference between multi-turn conversational agents and regular (1-turn) conversational agents is that instead of dealing with an utterance in a context on the \emph{word-level}, multi-turn models deal with a dialogue on the \emph{utterance-level}, so that models can understand an utterance as a whole and focus on important \emph{utterances} rather than important \emph{words}. An example of important/unimportant utterances existing in the same context is given by Table~\ref{tab:intro}.

\begin{center}
    \begin{longtable}{c|m{6cm}}
    \caption{\centering An example of important utterances and unimportant utterances under the same context in the Ubuntu chatlog dataset \cite{ubuntu}. Unimportant utterances are marked in {\color{red}red}.}\label{tab:intro}\\
    \hline
         \textbf{User} & \textbf{Utterances}\\
         \hline
         Taru & \color{red}Haha sucker.\\
         Kuja & \color{red}?\\
         Taru & Anyways, you made the changes right?\\
         Kuja & Yes.\\
         Taru & Then from the terminal type: sudo apt-get update\\
         Kuja & I did.\\
    \hline
    \end{longtable}
\end{center}

\vspace{-25pt}

In this example, the first two utterances (``Haha sucker.'' and ``?'') are unimportant utterances that are irrelevant to the main topic of the context. Human dialogues naturally contain many of these unimportant utterances. These utterances do not distract humans from understanding the main idea of the context, since humans can easily ignore them and focus instead on important utterances; however, a model usually lacks this capability and can be distracted by these utterances, resulting in a lower performance in generating {\em relevant} responses to the main topic of a context. Therefore, it is crucial that a multi-turn model can decide which utterances in the context are important and which are unimportant, and distribute its attention accordingly. In this paper, we define the research topic as \textbf{context attention distribution}, which denotes how much attention is distributed respectively to important and unimportant utterances in a context. A model with a good performance on context attention distribution should pay more attention to important utterances and less attention to unimportant utterances.

% multi-turn models study integration, attention mechanism, module construction, etc. 

Recent work lacks a measurement for the performance of multi-turn modelling. Common metrics rely on general evaluation metrics such as BLEU \cite{BLEU}, which measures the quality of generated responses. These metrics cannot directly describe a model's ability on dealing with multi-turn contexts, since the quality of generated responses is influenced by many aspects. Better performance in dealing with multi-turn context may result in better general performance; however, a better general performance does not necessariy mean that the model has a better ability on dealing with multi-turn contexts. Thus, as a supplementary to general evaluation metrics like BLEU, we propose a metric that measures a conversational agent's performance on context attention distribution, which is specifically designed for evaluating a model's performance on multi-turn modelling.

Since most multi-turn conversational agents have the attention mechanism and rely on it to distribute attention to different utterances in a context, we propose \textbf{distracting test} as the evaluation method to examine if a model pays more attention to the important utterances. The test adds unrelated utterances as distractions to the context of each dialogue and compares the attention scores of distracting utterances (i.e., unimportant utterances) and original utterances (i.e., important utterances). The ratio of the average attention score of distracting utterances and original utterances is defined as the distracting attention score ratio (\textbf{DAS ratio}). We use DAS ratio as the evaluation metric for a model's performance on context attention distribution. A model with good capability on context attention distribution should have higher scores on original utterances and lower scores on distracting utterances, thus a lower DAS ratio.

Furthermore, we propose a self-contained optimization strategy to improve a conversational agent's performance on context attention distribution. For each dialogue, we randomly pick some utterances from the training corpus outside the current dialogue as self-contained distractions, and insert them into the current dialogue with different levels of possibilities. The attention paid to these distractions is minimized during the training process through multi-task learning. With this optimization strategy, a model learns to distribute less attention to unimportant utterances and thus more attention to important utterances.

In this paper, we examine the following research questions: 1) How do existing multi-turn modelling structures perform on context attention distribution? 2) Can the proposed optimization strategy improve a model's performance on context attention distribution? 3) Which probability level is the best for inserting distractions in the proposed optimization strategy?

Our contributions are as follows:

\begin{enumerate}
    \item We deal with a less studied problem: evaluating and improving context attention distribution for multi-turn conversational agents.
    \item We propose a novel evaluation metric for multi-turn conversational agents: DAS ratio. It measures a model's performance on context attention distribution, i.e. the capability of distributing more attention to important utterances and less to unimportant ones.
    \item We propose an optimization strategy that minimizes the attention paid to self-contained distractions during the training process, and thus makes the model try to pay less attention to unimportant utterances. The strategy can easily be added and adapted to existing models.
\end{enumerate}

\vspace{-2pt}

Extensive experiments on $23$ model variants and $9$ distracting test sets show an overall improvement in the performance on context attention distribution for the proposed strategy. We will share our code for reproducibility.

Related work is introduced in Section 2. In Section 3, we introduce our base models and proposed methods. We show our experiments settings in Section 4 and results in Section 5. Finally, we give a conclusion in Section 6.

\section{Related Works}

\label{related_work}

Common evaluation metrics for conversational agents measure the similarity between the generated responses and the gold responses. Liu et al. \cite{evaluation-review} summarizes commonly used metrics: word overlap-based metrics (e.g. BLEU) and embedding-based metrics. Bruni et al. \cite{adversarial-evaluation} propose an adversarial evaluation method, which uses a classifier to distinguish human responses from generated responses. Lowe et al. \cite{lowe-evaluation} propose a model that simulates human scoring for generated responses. Zemlyanskiy et al. \cite{persona-evaluation} examine the quality of generated responses in a different direction: how much information the speakers exchange with each other. Recently, Li et al. \cite{dialogflow} propose a metric that evaluates the human-likeness of the generated response by measuring the gap between the corresponding semantic influences. Different from the above, our proposed evaluation metric is based on the attention mechanism and is intended to measure a model's performance on context attention distribution.

Most generation-based conversational agents apply simple concatenation for multi-turn conversation modelling \cite{knowledge2020,dialogpt}, which regards a multi-turn context as a 1-turn utterance. Some works try to model multi-turn conversations through the hierarchical structure: Serban et al. \cite{iulian1,iulian2} first introduce the hierarchical structure to dialogue models. Tian et al. \cite{context-review} evaluate different methods for integrating context utterances in hierarchical structures. Zhang et al. \cite{dual-attention-context} further evaluate the effectiveness of static and dynamic attention mechanism. Gu et al. \cite{dialogbert} apply a similar hierarchical structure on Transformer, and propose masked utterance regression and distributed utterance order ranking for the training objectives. Different from hierarchical models, Li et al. \cite{dialogflow} encode each utterance with a special token [$C$] and apply a flow module to train the model to predict the next [$C$]; then they use semantic influence (the difference of the predicted and original tokens) to support generation. In our paper, instead of modelling the relations of inter-context utterances as \cite{dialogbert} or the dialogue flow as \cite{dialogflow}, our optimization strategy improves multi-turn modelling by distinguishing important/unimportant utterances directly on the attention mechanism.

\section{Methods}
\label{methodology}

Our proposed evaluation metric and optimization strategy can work on attention mechanisms including Transformers. In this paper, we choose an LSTM Seq2Seq model with attention mechanism \cite{lstm,s2s,attention} as the base model, since most hierarchical structured multi-turn conversational agents are based on LSTM \cite{iulian1,iulian2,context-review,dual-attention-context} while few are based on Transformers. 

% Transformer-based model DialoFlow \cite{dialogflow} has an interesting structure with semantic influences instead of attention mechanism for supporting generation, so we still stick to the LSTM Seq2Seq model with attention.

The basic task of generation-based conversational agents is to predict the next token given all the past and current tokens from the context and response, and to make the predicted response as similar to the original response as possible. Formally, the probability of response $Y$ given context $X$ is predicted as:

\begin{equation}\textstyle P(Y|X)=\prod_{t=1}^{n}p(y_t|y_1,\ldots,y_{t-1},X)\text{,}\end{equation}

where $X=x_1,\ldots,x_m$ and $Y=y_1,\ldots,y_n$ are a context-response pair.

\subsection{LSTM Seq2Seq Model with Attention}\label{lstm}

We simplify an LSTM unit as $LSTM$, and we denote the attention version of an LSTM with an asterisk ($LSTM^*$). They are well introduced in previous work \cite{li-persona}. We calculate the hidden vector $h_t$ at step $t$ as:

\begin{equation}h_t = LSTM^*(h_{t-1},E(z_t),c_{t-1})\text{,}\end{equation}

where $h_{t-1}\in\R^{dim}$ is the hidden vector at step $t\text{-}1$, $dim$ is the dimensionality of hidden vectors, and $E(z_t)$ is the word embedding for token $z_t\in\{x_1,...,x_m,y_1,...,y_{n-1}\}$. $c_{t-1}$ is the context vector at step $t\text{-}1$, and it is input to the next step $t$ \emph{only in the decoder}. Each $h_t$ and $c_{t}$ of the current step $t$ are combined through a linear layer and an activation to predict the next token.

\subsection{Attention Mechanism \& Utterance Integration (UI)}\label{attention}

\label{model}
\begin{figure}[!ht]
    \centering
    \includegraphics[width=\textwidth,height=5.5cm]{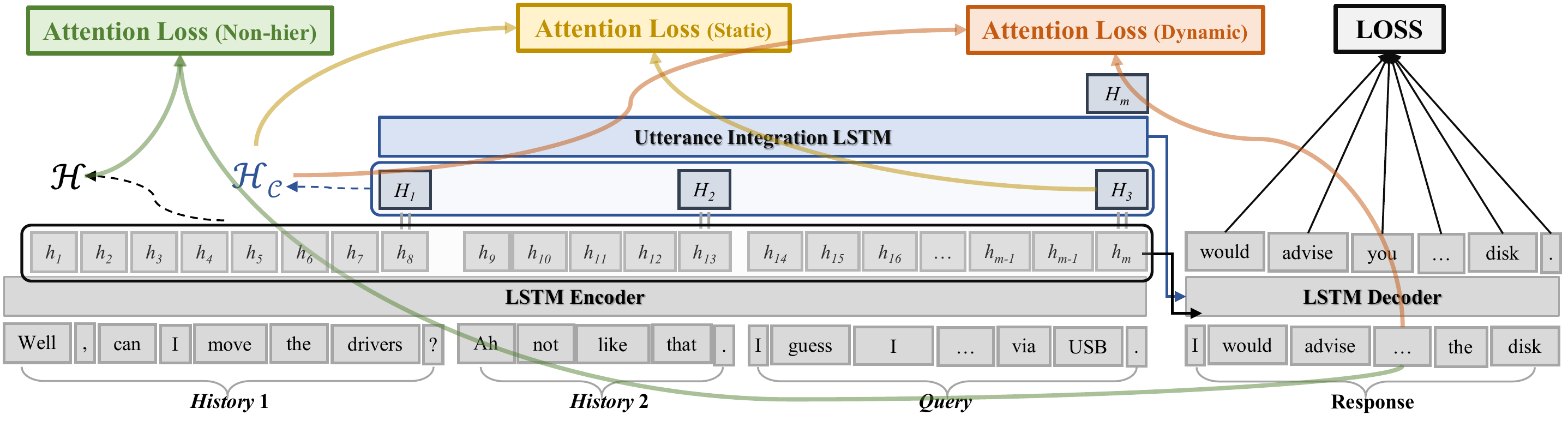}
    \caption{Structure of non-hierarchical, static and dynamic attention loss.}
    \label{fig}
\end{figure}

We examine both non-hierarchical and hierarchical structures. For hierarchical structures, following \cite{dual-attention-context}, we develop two attention mechanisms: static and dynamic. Following \cite{context-review}, we develop models that are both with and without utterance integration LSTM units.

For the non-hierarchical structured model, there are no hidden vectors for utterances. All hidden vectors of tokens in the encoder are concatenated and used in the attention mechanism. Denoting the concatenated vector $\mathcal{H}=[h_1,h_2,...,h_m]$, we calculate the context vector $c_t$ for each decoding step $t$ as: 

\begin{equation}c_t=\mathcal{H}\cdot(\text{softmax}(\mathcal{H}^\top\cdot h_t))\ \text{.}\end{equation}

For the hierarchical models, we use the hidden vector of each utterance's last token as the hidden vector of the utterance, and we discard the hidden vectors for the other tokens. Thus, compared to the non-hierarchical structured model, we have much fewer hidden vectors from the encoder.

The context vector of static attention mechanism is calculated based on the utterance-level concatenated vector and the hidden vector of the last utterance in the context. Denoting the hidden vector of $k$th utterance as $H_k$, and the hidden vector of the last utterance in the context as $H_q$, we have the context's concatenated vector $\mathcal{H}_\mathcal{C}=[H_1,H_2,...,H_q]$. We calculate the context vector $c_t$ for static attention mechanism as: 

\begin{equation}c_t=\mathcal{H}_\mathcal{C}\cdot(\text{softmax}(\mathcal{H}_\mathcal{C}^\top\cdot H_q))\text{,}\end{equation}

where it is easy to see that the static context vector remains unchanged by the decoder.

The context vector of dynamic attention mechanism is calculated based on the utterance-level concatenated vector and the hidden vector of each token in the decoding step. We calculate the context vector $c_t$ for dynamic attention mechanism as: 

\begin{equation}c_t=\mathcal{H}_\mathcal{C}\cdot(\text{softmax}(\mathcal{H}_\mathcal{C}^\top\cdot h_t))\ \text{.}\end{equation}

Compared to the static attention mechanism, the context vector $c_t$ varies at each decoding step.

Finally, with the utterance integration LSTM unit, we calculate $H_m$ from $H_1$, $H_2$, ... $H_q$:

\begin{equation}H_m = LSTM(H_1,H_2,...,H_q)\text{.}\end{equation}

For models with utterance integration (UI), $H_m$ is input to the first step of the decoder, while for models without UI, regular $h_m$ is input instead.

\subsection{Distracting Test \& Attention Score (AS)}

We examine if a multi-turn conversational agent distributes more attention to important utterances through the distracting test and attention scores.

In the distracting test, for each dialogue before the end of the context, we insert several utterances that are irrelevant to the main idea of the dialogue as distractions. These utterances are named {\em distracting utterances}, and they can be randomly picked utterances from the training corpus (\textbf{random}), be formed by frequent words from the training corpus (\textbf{frequent}), or be formed by rare words from the training corpus (\textbf{rare}). We compare the attention scores of the distracting utterances with the attention scores of the original utterances. A well-performing model should distribute less attention to the distracting utterances while more attention to the original utterances. For an utterance $H_k$, the corresponding attention score $\text{AS}(H_k)$ is calculated as:

\begin{equation}\text{AS}(H_k)=\begin{cases}\cfrac{m}{q}\cdot\mathtt{mean}_t\left(\cfrac{\sum_{h_i\in H_k}^{}\exp(h_i^\top\cdot h_t)}{\sum_{i=1}^{m}\exp(h_i^\top\cdot h_t)}\right)&\text{Non-hierarchical}\\\cfrac{q\cdot\exp(H_k^\top\cdot H_q)}{\sum_{k=1}^{q}\exp(H_k^\top\cdot H_q)}&\text{Static attention}\\ \mathtt{mean}_t\left(\cfrac{q\cdot\exp(H_k^\top\cdot h_t)}{\sum_{k=1}^{q}\exp(H_k^\top\cdot h_t)}\right)&\text{Dynamic attention}\\\end{cases}\text{.}\end{equation}

$h_i$ denotes hidden vectors from the encoding steps and $h_t$ denotes hidden vectors from the decoding steps. $m$ is the number of tokens in a context, and $q$ denotes the number of utterances in a context. Note that for non-hierarchical models we multiply by an $m$ in each $\text{AS}(H_k)$ to avoid bias caused by the total number of tokens in different contexts. Similarly for hierarchical models, we multiply by a $q$ in each $\text{AS}(H_k)$ to avoid bias caused by the number of total utterances in different contexts. As a result, for an utterance $H_q$, $\text{AS}(H_q)$ will be $100\%$ (or approximately $100\%$ for non-hierarchical models) if the model assigns $H_q$ an about average attention score among all utterances.

We denote the last utterance in a context as \emph{Query} and the rest of utterances in the context as \emph{History}. Since different models have different scalars on attention scores, we calculate the average AS for all distracting utterances and all \emph{History} in each dialogue, and use the ratio of them for evaluation. This ratio is denoted as distracting attention score ratio (\textbf{DAS ratio}), which measures a model's ability on context attention distribution:

\begin{equation}
    \text{DAS ratio} =\mathtt{mean}_{d\in D}\left({ \frac{\mathtt{mean}(\text{AS}(H_{\text{Distraction}}))}{\mathtt{mean}(\text{AS}(H_{\text{\em History}}))}}\right)
\text{,}\end{equation}

where $d$ means a single dialogue, and $D$ denotes all dialogues in a test set. $H_{\text{Distraction}}$ denotes distracting utterances, and $H_{\text{\em History}}$ denotes utterances in {\em History}.

\subsection{Optimization with Self-Contained Distractions on Attention Mechanism}

% The main idea of our optimization strategy A model is given labels on which utterances are distracting utterances, and it learns to assign lower attention scores to these utterances through multi-task learning. Thus, it learns to distribute less attention to unimportant utterances and more attention to important utterances in a context.

To train a conversational model to distribute more attention to important and less attention to unimportant utterances, we propose the following optimization strategy: 1) For each dialogue, we select some random utterances from other dialogues in the training corpus as \textbf{self-contained distractions}. We decide whether to insert these distractions into the current dialogue or not stochastically by a probability level. We denote the probability level as the training inserting probability. The locations of inserting distractions are randomly decided, while the locations are always before {\em Query} (the last utterance of the context). 2) We create a bitmask $M$ to track whether an utterance is original ($0$) or distracting ($1$). During the training period, the model uses the bitmask to calculate the attention loss $\mathcal{L}^t_{\mathrm{attention}}$, which is summed up with the loss from the response generator. For each decoding step $t$, the attention loss is calculated as:

\begin{equation}\mathcal{L}^t_{attention}={\begin{cases}\mathtt{MSE}(\text{softmax}(\mathcal{H}^\top\cdot h_t)\circ M,0)&\text{Non-hierarchical}\\ \mathtt{MSE}(\text{softmax}(\mathcal{H}^\top_\mathcal{C}\cdot H_q)\circ M,0)&\text{Static attention}\\ \mathtt{MSE}(\text{softmax}(\mathcal{H}^\top_\mathcal{C}\cdot h_t)\circ M,0) &\text{Dynamic attention}\\\end{cases}}\label{equa}\end{equation}

where $\circ$ means Hadamard product, or elementwise multiplication. As shown in Equation (\ref{equa}), our goal is to minimize the attention assigned to all the self-contained distractions. During the distracting test, no bitmask is offered to the model. The illustration of attention loss on both non-hierarchical and hierarchical models is shown in Figure~\ref{fig}.

\section{Experiments}
\label{experiment}

In this section, we introduce the setups of the experiment.

\subsection{Dataset}

We use the Ubuntu chatlogs dataset \cite{ubuntu} as the training and testing corpus, which contains dialogues about solving technical problems of Ubuntu. We choose this dataset because the dialogues have both technical topics and casual chats, meaning that it is easier to distinguish important/unimportant utterances than datasets whose topics are consistent. We use about 0.48$M$ dialogues for training, 20$K$ dialogues for validation, and 10$K$ dialogues for testing. These are the original settings of the Ubuntu chatlogs dataset. We removed all single-turn dialogues.

\subsection{Training}

Our methods are built on an LSTM Seq2Seq model with attention mechanism. We used Pytorch \cite{pytorch} for implementation. The LSTM model has 4 layers and the dimension is 512. The training procedure was with a batch size of 256, a learning rate of 1.0, and a gradient clip threshold of 5. The vocabulary size is 25000 and the dropout rate is 0.2. The learning rate is halved when the perplexity stops dropping, and the training is stopped when the model converges.

\subsection{Examined Models}

We examine our proposed evaluation metric on 5 models: non-hierarchical LSTM (Non-hier), static attention without utterance integration LSTM unit (Static), static attention with utterance integration LSTM unit (StaticUI), dynamic attention without utterance integration LSTM unit (Dynamic), and dynamic attention with utterance integration LSTM unit (DynamicUI). In addition, we examine our proposed optimization strategy on these 5 models with 3 training inserting probabilities--0.5, 0.7, and 1.0. Models with a training inserting probability of $0$ are regarded as baselines. For comparison, we pick the best overall model and train the model with self-contained distractions but without training on the attention loss (Non-atten-loss), i.e. the model does not know which utterances are distractions. In total, we train and evaluate 23 model variants.

\subsection{Evaluation}

\begin{table}[!ht]
    \centering
    \caption{Examples of distracting test sets. Distracting utterances are marked {\color{red}red}.}
    \label{tab:example}
    
    \begin{tabular}{m{1.3cm}|m{3.9cm}|m{3.9cm}|m{3.9cm}}
    \toprule
    &\multicolumn{1}{c|}{\textbf{Random: 0.5}} & \multicolumn{1}{c|}{\textbf{Random: 0.7}} & \multicolumn{1}{c}{\textbf{Random: 1.0}}\\
    \hline
    \multirow{4}{*}{\em History} & \multicolumn{1}{c|}{$\backslash$} & Well, can I move the drives? & \color{red}Yes.\\
    \cline{2-4}
    & \color{red} Or kill all speedlink. & \color{red} Anyways, you made the changes right? & Well, can I move the drives?\\
    \cline{2-4}
    & Well, can I move the drives? & Ah not like that. & \color{red}Then from the terminal type: sudo apt-get update. \\
    \cline{2-4}
    & Ah not like that. & \color{red} I did. & Ah not like that.\\
    \hline\hline
    &\multicolumn{1}{c|}{\textbf{Frequent: Begin}} & \multicolumn{1}{c|}{\textbf{Frequent: Middle}} & \multicolumn{1}{c}{\textbf{Frequent: End}}\\
   \hline
    \multirow{4}{*}{\em History} & \color{red}Why should I help you? & Well, can I move the drives? & Well, can I move the drives?\\
    \cline{2-4}
    & \color{red}I have my right. & \color{red}Why should I help you? & Ah not like that. \\
    \cline{2-4}
    & Well, can I move the drives? & \color{red}I have my right. & \color{red}Why should I help you? \\
    \cline{2-4}
    & Ah not like that. &  Ah not like that. & \color{red}I have my right.\\
    \hline\hline
    &\multicolumn{1}{c|}{\textbf{Rare: Begin}} & \multicolumn{1}{c|}{\textbf{Rare: Middle}} & \multicolumn{1}{c}{\textbf{Rare: End}}\\
    \hline
    \multirow{4}{*}{\em History} & \color{red}Would you have lunch? & Well, can I move the drives? & Well, can I move the drives?\\
    \cline{2-4}
    & \color{red}I should have lunch. & \color{red}Would you have lunch? & Ah not like that. \\
    \cline{2-4}
    & Well, can I move the drives? & \color{red}I should have lunch. & \color{red}Would you have lunch? \\
    \cline{2-4}
    & Ah not like that. &  Ah not like that. & \color{red}I should have lunch.\\
    \hline\hline
    \em Query & \multicolumn{3}{c}{\textbf{I guess I could just get an enclosure and copy via USB.}}\\
    \hline
    Response & \multicolumn{3}{c}{\textbf{I would advise you to get the disk.}}\\
    \bottomrule
    \end{tabular}

\end{table}

For the distracting test, we set the number of distracting utterances for each dialogue to $2$. We chose $2$ to make the distracting utterances a complete turn and to make the number of distracting utterances the minimum, since dialogues from the corpus normally have only $4$ to $8$ utterances in the contexts. We have $3$ distracting test sets. 1) Random distracting test set: distracting utterances in this test set are randomly picked from the training corpus (outside the current dialogue), and they are randomly picked in every evaluation step, which means that there is no pre-prepared random distracting test set. 2) Frequent distracting test set: distracting utterances in this test set are formed by frequent words in the training corpus, but these utterances do not appear in the training corpus. In our experiments, we use ``why should I help you'' and ``I have my right'' as examples of distracting utterances with frequent words. 3) Rare distracting test set: distracting utterances in this test set have words that are rare in the training corpus, and these utterances do not appear in the training corpus. In our experiments, we use ``would you have lunch?'' and ``I should have lunch'' as examples of distracting utterances with rare words.

In the distracting test, we insert distracting utterances into different locations. For 1) random, we insert utterances to a random location before {\em Query} in each context. Similar to the optimization strategy, we use different probability levels to decide whether a distracting utterance is to be inserted or not. We denote these as testing inserting probability. In our experiments, we set the probability levels to be 0.5, 0.7, and 1.0. We expect the model to perform stably on all different probability levels. For 2) frequent and 3) rare, we have three kinds of inserting locations: at the beginning of a context (marked as Begin), in the middle of the context (marked as Middle), and at the end of the context (before {\em Query} and after {\em History}, marked as End). In total, we have 9 test sets for evaluation. See Table \ref{tab:example} for the example of each test set.

\section{Results and Discussions}
\label{result}

Table~\ref{table:result} illustrates the main results on DAS ratios. It shows the DAS ratios of $23$ trained model variants on $9$ distracting test sets. Figure \ref{fig_result} shows the DAS ratios of $3$ example model variants (StaticUI with training inserting probability of 0.0 as the baseline, Non-atten-loss StaticUI with training inserting probability of 0.7, and StaticUI with training inserting probability of 0.7) on $9$ distracting test sets. Table~\ref{table:random}, Table~\ref{table:frequent} and Table~\ref{table:rare} show the detailed results on average Attention Score (average AS) of distracting utterances and average AS of {\em History}.

In Table~\ref{table:result}, we show the perplexity and {\em History}'s average AS of each model on the non-distracted test set under the ``Original'' column. Since perplexity scores on the distracting test sets are similar, we show the perplexity scores on the non-distracted test set only. We show the DAS ratios of each model on each of the distracting test sets under the ``DAS ratio for distracting test set'' column. A lower DAS ratio means that a model distributes less attention to distracting utterances (unimportant utterances) and more attention to the original utterances in {\em History} (important utterances), from which it can be inferred that the model has better performance on context attention distribution. Both perplexity and DAS ratio are the lower, the better.

\subsection{Perplexity and Average AS on Non-Distracted Test Set}

Perplexity scores are shown in the ``Perp.'' column, under the ``Original'' column in Table \ref{table:result}. Perplexity scores of the examined $23$ models are similar; the Static models trained with our proposed optimization strategy and a higher training inserting probability level achieves slightly better performance than other models.

Average AS are shown in the ``Avg.'' column, under the ``Original'' column in Table \ref{table:result}. The average AS of {\em History} tells about a model's attention distribution for {\em History} and {\em Query}. A higher score indicates that less attention is distributed to {\em Query}. Recall that AS of an utterance is $100\%$ (or approximately $100\%$ for non-hierarchical models) if the utterance is paid about average attention among the dialogue. Overall, the models distribute attention of lower than average to {\em History}, especially for models with static attention (i.e. the Static model and StaticUI model), which distribute more attention to {\em Query} than non-hierarchical models and models with dynamic attention. This is apparent from the structure of static attention. We also show the results of a StaticUI model without training on the attention loss (Non-atten-loss StaticUI model) as a comparison. The StaticUI model trained with our optimization strategy distributes more attention to {\em query} than the Non-atten-loss StaticUI model. This is because the optimization strategy decreases the model's attention distributed to distracting utterances in {\em History}, thus decreasing the overall attention distributed to {\em History}.

\subsection{Distracting Test: Random}

Results of the random distracting test with different testing inserting probabilities ($0.5$, $0.7$, and $1.0$) are shown in the ``Random'' column in Table \ref{table:result}. Models with training inserting probabilities of $0.0$ (shown in the row where ``Prob'' is $0.0$) are baseline models to which our proposed optimization strategy is not applied. In general, our proposed optimization strategy with training inserting probabilities of $0.5$ or $0.7$ achieves better performance on DAS ratios (i.e. the models achieve lower DAS ratios) on random distracting test sets of all $3$ testing inserting probabilities. The Static model and the DynamicUI model achieves the best performance with a training inserting probability of $0.5$, while the Non-hier model, the StaticUI model and the Dynamic model achieve the best performance with a training inserting probability of $0.7$. A training inserting probability of $1.0$ leads to worse performance. One reason is that it assumes there must be some distracting utterances in a context, while that is not always the case.

% @{\hspace{0.2\tabcolsep}}
\newcommand{\resulta}[4]{#2 & #1 & #3 & #4}
\newcommand{\resultb}[3]{#2 & #1 & #3}
\newcommand{\resultaa}[4]{\multirow{2}{*}{#2} & \multirow{2}{*}{#1} & \multirow{2}{*}{#3} & \multirow{2}{*}{#4}}
\newcommand{\resultbb}[3]{\multirow{2}{*}{#2} & \multirow{2}{*}{#1} & \multirow{2}{*}{#3}}
\newcommand{\result}[3]{#1 & #2 & #3}
\newcommand{\resultt}[3]{\multirow{2}{*}{#1} & \multirow{2}{*}{#2} & \multirow{2}{*}{#3}}

\begin{table}[ht!]
\centering

\caption{Results of perplexity (Perp.) and average AS of \emph{History} (Avg.) on the original test set (\%) are shown in the ``Original'' column. We also show results of DAS ratios on $9$ distracting test sets and $23$ model variants.}
	\label{table:result}
	
\begin{adjustbox}{angle=90}
	\begin{tabular}{c|l|cc|ccc|ccc|ccc}
% 	\begin{tabular}{@{\hspace{0.5\tabcolsep}}c|l|cc|ccc|ccc|ccc@{\hspace{0\tabcolsep}}}
	\toprule
	    \multicolumn{2}{c|}{Model} & \multicolumn{2}{c|}{Original} &\multicolumn{9}{c}{DAS ratio on distracting test sets}\\
	\hline
		\multirow{2}{*}{Prob} & \multirow{2}{*}{Structure} & \multirow{2}{*}{\footnotesize{Perp.}} & \multirow{2}{*}{\footnotesize{Avg.}} & \multicolumn{3}{c|}{Random} & \multicolumn{3}{c|}{Frequent} & \multicolumn{3}{c}{Rare}\\

		& & & & \result{\footnotesize{0.5}}{\footnotesize{0.7}}{\footnotesize{1.0}} & \result{\footnotesize{Begin}}{\footnotesize{Middle}}{\footnotesize{End}} & \result{\footnotesize{Begin}}{\footnotesize{Middle}}{\footnotesize{End}} \\
	\hline
		\multirow{5}{*}{$0.0$} & Non-hier & 43.2&91.3 & \result{0.93}{0.93}{0.93} & \result{0.75}{0.80}{0.84} & \result{0.80}{0.92}{1.01} \\
		& Static & 44.1&61.4 & \result{0.82}{0.82}{0.79} & \result{0.37}{0.80}{1.31} & \result{0.37}{0.77}{1.21} \\
		& StaticUI & 44.6&57.5 & \result{0.79}{0.76}{0.76} & \result{\cellcolor{r}\textbf{0.32}}{0.75}{1.32} & \result{0.30}{0.75}{1.22} \\
		& Dynamic & 45.4&81.4 & \result{0.89}{0.89}{0.88} & \result{0.65}{0.86}{1.02} & \result{0.66}{0.89}{1.06} \\
		& DynamicUI & 44.7&91.6 & \result{0.94}{0.94}{0.93} & \result{0.72}{0.84}{0.86} & \result{0.73}{0.93}{0.97} \\
	\hline
		\multirow{6}{*}{$0.5$} & Non-hier & 43.4&87.2 & \result{0.84}{0.83}{0.81} & \result{0.63}{0.74}{\cellcolor{r}\textbf{0.76}} & \result{0.69}{0.81}{0.86} \\
		& Static & 44.5&66.5 & \result{0.70}{0.69}{0.67} & \result{0.42}{0.78}{1.12} & \result{0.34}{0.71}{0.99} \\
		& StaticUI & 44.3&47.7 & \result{0.74}{0.74}{0.70} & \result{0.39}{0.71}{1.08} & \result{0.40}{\cellcolor{r}\textbf{0.69}}{0.96} \\
		& Dynamic & 44.6&81.9 & \result{0.79}{0.78}{0.77} & \result{0.64}{0.74}{0.84} & \result{0.61}{0.77}{0.85} \\
		& DynamicUI & 43.9&86.7 & \result{0.82}{0.81}{0.80} & \result{0.60}{0.84}{0.87} & \result{0.61}{0.80}{0.83} \\
	\cline{2-13}
		& Non-atten-loss & \multirow{2}{*}{44.7}&\multirow{2}{*}{71.1} & \resultt{0.73}{0.73}{0.72} & \resultt{0.39}{0.68}{0.93} & \resultt{0.40}{0.80}{1.11} \\
		& StaticUI & & & && & && & && \\
	\hline
		\multirow{6}{*}{$0.7$} & Non-hier & 43.2&86.9 & \result{0.84}{0.82}{0.80} & \result{0.72}{0.82}{0.82} & \result{0.71}{0.85}{0.87} \\
		& Static & \cellcolor{r}\textbf{44.0}&57.6 & \result{0.73}{0.72}{0.69} & \result{0.40}{0.70}{1.08} & \result{0.41}{0.70}{0.98} \\
		& StaticUI & 44.9& 43.7 & \result{\cellcolor{r}\textbf{0.67}}{\cellcolor{r}\textbf{0.67}}{\cellcolor{r}\textbf{0.65}} & \result{0.36}{\cellcolor{r}\textbf{0.66}}{1.02} & \result{0.36}{0.70}{0.99} \\
		& Dynamic & 44.3&82.0 & \result{0.76}{0.75}{0.73} & \result{0.58}{0.71}{0.86} & \result{0.58}{0.73}{0.83} \\
		& DynamicUI & 44.8&85.3 & \result{0.93}{0.93}{0.93} & \result{0.45}{0.78}{0.80} & \result{0.60}{0.80}{\cellcolor{r}\textbf{0.81}} \\
	\cline{2-13}
		& Non-atten-loss & \multirow{2}{*}{44.1}&\multirow{2}{*}{55.4} & \resultt{0.72}{0.70}{0.69} & \resultt{0.45}{0.70}{0.98} & \resultt{0.43}{0.73}{0.97} \\
		& StaticUI & & & && & && & && \\
	\hline
		\multirow{6}{*}{$1.0$} & Non-hier & 47.3&95.9 & \result{0.91}{0.90}{0.90} & \result{0.84}{0.86}{0.85} & \result{0.85}{0.87}{0.88} \\
		& Static & \cellcolor{r}\textbf{44.0}&65.4 & \result{0.70}{0.70}{0.68} & \result{0.49}{0.74}{1.08} & \result{0.46}{0.71}{0.88} \\
		& StaticUI & 49.6&73.5 & \result{0.96}{0.95}{0.94} & \result{0.66}{0.86}{1.53} & \result{\cellcolor{r}\textbf{0.21}}{0.86}{1.50} \\
		& Dynamic & 44.7&88.8 & \result{0.79}{0.78}{0.77} & \result{0.63}{0.75}{0.82} & \result{0.65}{0.77}{0.82} \\
		& DynamicUI & 45.2&90.2 & \result{0.87}{0.86}{0.85} & \result{0.73}{0.81}{0.83} & \result{0.75}{0.88}{0.88} \\
	\cline{2-13}
		& Non-atten-loss & \multirow{2}{*}{44.1}&\multirow{2}{*}{76.5} & \resultt{0.72}{0.71}{0.69} & \resultt{0.49}{0.74}{0.98} & \resultt{0.49}{0.77}{0.98} \\
		& StaticUI & & & && & && & && \\
	\bottomrule

	\end{tabular}
\end{adjustbox}
	
\end{table}

\newpage

\begin{figure}[!ht]
    \centering
    \includegraphics[width=\textwidth,height=6cm]{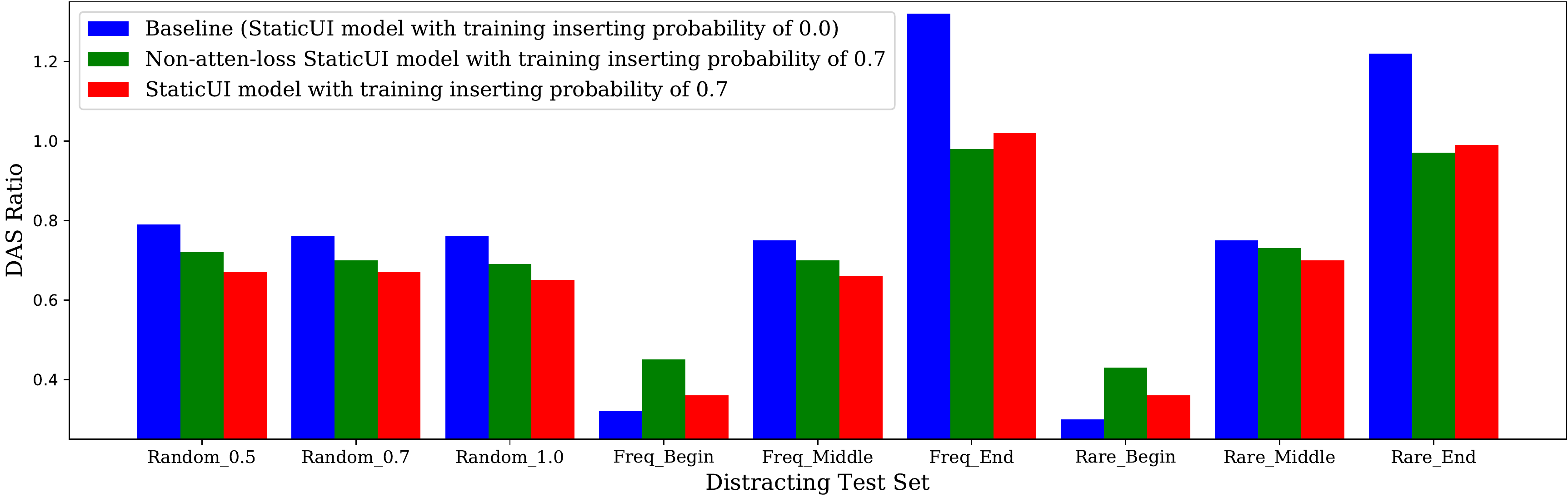}
    \caption{\centering DAS ratios of $3$ example model variants on $9$ distracting test sets. The lower the DAS ratio, the better the performance.}
    \label{fig_result}
\end{figure}

\vspace{-10pt}

The StaticUI model with a training inserting probability of $0.7$ achieves the best overall performance on DAS ratio. As shown in Figure \ref{fig_result}, on all the random distracting test sets (probabilities of $0.5$, $0.7$, and $1.0$), the StaticUI model is better than the baseline StaticUI model and the Non-atten-loss StaticUI model. The baseline model is not trained with any self-contained distractions (training inserting probability is $0.0$), and it gets the worst performance. The Non-atten-loss model is trained with self-contained distractions (with a training inserting probability of $0.7$) while not knowing which utterances are distractions, and it achieves a better performance than the baseline. The StaticUI model with a training inserting probability of $0.7$ is trained to minimize the attention loss of self-contained distractions and it achieves the best performance. Naturally since the optimization strategy minimizes the attention loss of distractions, the StaticUI model distributes less attention to {\em History} and more attention to {\em Query} (refer to the ``Avg'' column in Appendix \ref{table:random} for more details); nevertheless, a lower DAS ratio shows that the model distributes even less attention to the distracting utterances compared to the original utterances in {\em History}.

Note that even if both our proposed strategy and the random distracting test use the same trick: insert random distracting utterances among original utterances in {\em History}, the random utterances inserted in the distracting test are different from those inserted in the training process, thus it is difficult for the test to be biased in favor of models with our proposed strategy. It is apparent that less attention is distributed to {\em History}, while DAS ratio calculates the ratio between the distracting utterances and the original utterances in {\em History}, so it shows the attention distributed to the distracting utterances regardless of the total attention distributed to {\em History}. Moreover, we adopt three testing inserting probability levels to ensure stable evaluation results for each model.
% so that models trained with our optimization strategy naturally suffer from these dialogues without distracting utterances.

\subsection{Distracting Test: Frequent and Rare}

Results of the frequent and the rare distracting test are shown in the ``Frequent'' and ``Rare'' columns in Table \ref{table:result}. Different from the random distracting test, the inserting locations of these two tests are decided manually. As a nature of LSTM model, all models distribute more attention to utterances near {\em Query} and less attention to utterances far away from {\em Query}, as can be seen in Table \ref{table:result} and Figure \ref{fig_result} that DAS ratios are higher for End test set (near {\em Query}) and lower for Begin test set (far away from {\em Query}). Since the results on Begin and End test sets are biased by the structure of LSTM, we mainly analyze the results on Middle test sets.

For the Middle test sets of both the frequent and rare distracting test, the best models are still those trained with our proposed optimization strategy. StaticUI models with training inserting probabilities of $0.5$ and $0.7$ achieve the best performance (lowest DAS ratios) on the Frequent Middle and Rare Middle test sets. The Non-atten-loss models can be better than the models trained with a wrong training inserting probability. Telling from similar DAS ratios, the frequent distracting test set is as difficult for the trained models to distinguish as the rare distracting test set, although for humans, the rare distracting utterances are much easier to distinguish than the frequent ones.

\begin{table}[!ht]
\centering
	\caption{Results of perplexity (Perp.) and average AS of {\em History} (Avg.) on the original test set (\%) are shown in the ``Original'' column. Besides, we show the results on the random distracting test of: DAS ratio, average AS of distracting utterances (DAS) (\%), and average AS of original utterances in {\em History} (Avg.) (\%).}
	\label{table:random}
\begin{adjustbox}{angle=90}

	\begin{tabular}{c|l|cc|ccc|ccc|ccc@{\hspace{0\tabcolsep}}}
	\toprule
	    \multicolumn{2}{c|}{Model} & \multicolumn{2}{c|}{Original} &\multicolumn{9}{c}{Distracting Test Set}\\
	\hline
		\multirow{2}{*}{Probability} & \multirow{2}{*}{Structure} & \multirow{2}{*}{Perp.} & \multirow{2}{*}{Avg.} & \multicolumn{3}{c|}{Random $0.5$} & \multicolumn{3}{c|}{Random $0.7$} & \multicolumn{3}{c}{Random $1.0$}\\

		& & & & \resultb{DAS}{DAS ratio}{Avg.} & \resultb{DAS}{DAS ratio}{Avg.} & \resultb{DAS}{DAS ratio}{Avg.} \\
	\hline
		\multirow{5}{*}{$0.0$} & Non-hier & 43.2&91.3 & \resultb{86.8}{0.93}{93.1} & \resultb{87.1}{0.93}{93.8} & \resultb{87.6}{0.93}{94.5} \\
		& Static & 44.1&61.4 & \resultb{52.6}{0.82}{64.5} & \resultb{53.5}{0.82}{65.6} & \resultb{53.4}{0.79}{67.6} \\
		& StaticUI & 44.6&57.5 & \resultb{47.4}{0.79}{60.2} & \resultb{46.3}{0.76}{61.3} & \resultb{47.7}{0.76}{62.7} \\
		& Dynamic & 45.4&81.4 & \resultb{74.9}{0.89}{84.4} & \resultb{75.6}{0.89}{85.2} & \resultb{76.2}{0.88}{86.6} \\
		& DynamicUI & 44.7&91.6 & \resultb{87.5}{0.94}{93.4} & \resultb{88.0}{0.94}{93.7} & \resultb{88.2}{0.93}{94.5} \\
	\hline
		\multirow{6}{*}{$0.5$} & Non-hier & 43.4&87.2 & \resultb{77.2}{0.84}{91.6} & \resultb{77.1}{0.83}{93.3} & \resultb{77.0}{0.81}{95.5} \\
		& Static & 44.5&66.5 & \resultb{50.3}{0.70}{71.5} & \resultb{50.5}{0.69}{73.5} & \resultb{51.1}{0.67}{76.5} \\
		& StaticUI & 44.3&47.7 & \resultb{38.1}{0.74}{51.2} & \resultb{39.2}{0.74}{53.1} & \resultb{39.1}{0.70}{55.5} \\
		& Dynamic & 44.6&81.9 & \resultb{68.3}{0.79}{86.6} & \resultb{69.1}{0.78}{88.2} & \resultb{69.4}{0.77}{90.8} \\
		& DynamicUI & 43.9&86.7 & \resultb{74.5}{0.82}{91.1} & \resultb{75.2}{0.81}{92.5} & \resultb{75.8}{0.80}{94.8} \\
	\cline{2-13}
		& Non-labled & \multirow{2}{*}{44.7}&\multirow{2}{*}{71.1} & \resultbb{55.5}{0.73}{75.6} & \resultbb{56.6}{0.73}{77.1} & \resultbb{57.4}{0.72}{79.6} \\
		& StaticUI & & & && & && & && \\
	\hline
		\multirow{6}{*}{$0.7$} & Non-hier & 43.2&86.9 & \resultb{76.5}{0.84}{91.3} & \resultb{75.9}{0.82}{93.1} & \resultb{75.9}{0.80}{95.4} \\
		& Static & \cellcolor{r}\textbf{44.0}&57.6 & \resultb{45.5}{0.73}{62.2} & \resultb{45.8}{0.72}{64.0} & \resultb{46.4}{0.69}{66.9} \\
		& StaticUI & 44.9&\cellcolor{r}\textbf{43.7} & \resultb{\cellcolor{r}\textbf{32.4}}{\cellcolor{r}\textbf{0.67}}{\cellcolor{r}\textbf{48.1}} & \resultb{\cellcolor{r}\textbf{33.2}}{\cellcolor{r}\textbf{0.67}}{\cellcolor{r}\textbf{49.9}} & \resultb{\cellcolor{r}\textbf{34.0}}{\cellcolor{r}\textbf{0.65}}{\cellcolor{r}\textbf{52.1}} \\
		& Dynamic & 44.3&82.0 & \resultb{66.3}{0.76}{87.2} & \resultb{66.6}{0.75}{89.1} & \resultb{67.2}{0.73}{91.8} \\
		& DynamicUI & 44.8&85.3 & \resultb{86.8}{0.93}{93.1} & \resultb{87.1}{0.93}{93.8} & \resultb{87.6}{0.93}{94.5} \\
	\cline{2-13}
		& Non-labled & \multirow{2}{*}{44.1}&\multirow{2}{*}{55.4} & \resultbb{43.3}{0.72}{59.9} & \resultbb{43.4}{0.70}{62.0} & \resultbb{44.3}{0.69}{64.4} \\
		& StaticUI & & & && & && & && \\
	\hline
		\multirow{6}{*}{$1.0$} & Non-hier & 47.3&95.9 & \resultb{88.7}{0.91}{98.0} & \resultb{89.3}{0.90}{98.7} & \resultb{89.5}{0.90}{99.9} \\
		& Static & \cellcolor{r}\textbf{44.0}&65.4 & \resultb{49.7}{0.70}{71.1} & \resultb{51.3}{0.70}{73.1} & \resultb{51.8}{0.68}{76.4} \\
		& StaticUI & 49.6&73.5 & \resultb{74.8}{0.96}{77.8} & \resultb{75.2}{0.95}{79.4} & \resultb{76.7}{0.94}{81.2} \\
		& Dynamic & 44.7&88.8 & \resultb{74.2}{0.79}{93.4} & \resultb{74.4}{0.78}{95.2} & \resultb{75.4}{0.77}{97.4} \\
		& DynamicUI & 45.2&90.2 & \resultb{81.3}{0.87}{93.6} & \resultb{81.5}{0.86}{94.9} & \resultb{81.9}{0.85}{96.5} \\
	\cline{2-13}
		& Non-labled & \multirow{2}{*}{44.1}&\multirow{2}{*}{76.5} & \resultbb{59.5}{0.72}{82.1} & \resultbb{59.7}{0.71}{84.4} & \resultbb{60.3}{0.69}{87.6} \\
		& StaticUI & & & && & && & && \\
	\bottomrule

	\end{tabular}
\end{adjustbox}
\end{table}

\newpage

\begin{table}[!ht]
\centering
	\caption{Results on the frequent distracting test of: DAS ratio, average AS of distracting utterances (DAS) (\%), average AS of original utterances in {\em History} (Avg.) (\%), and AS of the first/last utterance in {\em History} (\%).}
	\label{table:frequent}
\begin{adjustbox}{angle=90}
	\begin{tabular}{c|c|cccc|ccc|cccc@{\hspace{0\tabcolsep}}}
	% \toprule
	    \toprule
	    \multicolumn{2}{c|}{Model} & \multicolumn{11}{c}{Distracting Test Set}\\
	    \hline
		\multirow{2}{*}{Probability} & \multirow{2}{*}{Structure} & \multicolumn{4}{c|}{Frequent: Begin} & \multicolumn{3}{c|}{Frequent: Middle} & \multicolumn{4}{c}{Frequent: End}\\

		& & \resulta{DAS}{DAS ratio}{Avg.}{1st} & \resultb{DAS}{DAS ratio}{Avg.} &
		\resulta{DAS}{DAS ratio}{Avg.}{{Last}}\\
	\hline
		\multirow{5}{*}{$0.0$} & Non-hier & \resulta{70.8}{0.75}{94.1}{82.3} & \resultb{74.7}{0.80}{93.2} & \resulta{78.0}{0.84}{92.7}{98.4} \\
		& Static & \resulta{29.3}{0.37}{79.5}{47.9} & \resultb{56.7}{0.80}{71.2} & \resulta{82.0}{1.31}{62.4}{72.6} \\
		& StaticUI & \resulta{\cellcolor{r}\textbf{24.3}}{\cellcolor{r}\textbf{0.32}}{\cellcolor{r}\textbf{76.1}}{\cellcolor{r}\textbf{42.2}} & \resultb{50.6}{0.75}{67.7} & \resulta{77.2}{1.32}{58.7}{69.5} \\
		& Dynamic & \resulta{60.9}{0.65}{93.3}{72.7} & \resultb{75.5}{0.86}{88.0} & \resulta{87.5}{1.02}{85.1}{88.8} \\
		& DynamicUI & \resulta{72.2}{0.72}{100.7}{81.0} & \resultb{81.6}{0.84}{97.0} & \resulta{84.1}{0.86}{98.2}{98.8} \\
	\hline
		\multirow{6}{*}{$0.5$} & Non-hier & \resulta{57.3}{0.63}{91.0}{84.0} & \resultb{66.4}{0.74}{89.7} & \resulta{68.7}{\cellcolor{r}\textbf{0.76}}{90.8}{99.5} \\
		& Static & \resulta{35.0}{0.42}{84.5}{47.4} & \resultb{59.8}{0.78}{77.0} & \resulta{80.3}{1.12}{71.5}{81.7} \\
		& StaticUI & \resulta{24.6}{0.39}{62.9}{44.3} & \resultb{41.8}{0.71}{58.9} & \resulta{\cellcolor{r}\textbf{54.7}}{1.08}{\cellcolor{r}\textbf{50.6}}{\cellcolor{r}\textbf{55.2}} \\
		& Dynamic & \resulta{60.4}{0.64}{94.6}{76.7} & \resultb{68.2}{0.74}{92.1} & \resulta{76.2}{0.84}{91.1}{94.4} \\
		& DynamicUI & \resulta{60.1}{0.60}{100.4}{77.7} & \resultb{78.4}{0.84}{92.9} & \resulta{82.3}{0.87}{94.2}{93.0} \\
	\cline{2-13}
		& Non-labled & \resultaa{35.1}{0.39}{90.3}{53.7} & \resultbb{56.1}{0.68}{82.7} & \resultaa{72.6}{0.93}{78.4}{91.2} \\
		& StaticUI & &&& & && & &&& \\
	\hline
		\multirow{6}{*}{$0.7$} & Non-hier & \resulta{64.7}{0.72}{90.1}{83.1} & \resultb{72.6}{0.82}{88.8} & \resulta{73.4}{0.82}{89.4}{97.1} \\
		& Static & \resulta{30.3}{0.40}{75.2}{48.3} & \resultb{48.7}{0.70}{69.3} & \resulta{68.1}{1.08}{62.9}{69.2} \\
		& StaticUI & \resulta{21.0}{0.36}{57.6}{37.4} &\resultb{\cellcolor{r}\textbf{36.0}}{\cellcolor{r}\textbf{0.66}}{\cellcolor{r}\textbf{54.7}} &\resulta{51.7}{1.02}{50.6}{56.4} \\
		& Dynamic & \resulta{56.3}{0.58}{96.8}{73.9} &\resultb{66.4}{0.71}{93.4} &\resulta{76.0}{0.86}{88.8}{91.8} \\
		& DynamicUI & \resulta{44.4}{0.45}{98.8}{76.1} &\resultb{73.2}{0.78}{93.8} &\resulta{75.8}{0.80}{95.2}{95.0} \\
	\cline{2-13}
		& Non-labled & \resultaa{31.7}{0.45}{70.6}{51.0} &\resultbb{46.6}{0.70}{66.6} &\resultaa{60.2}{0.98}{61.2}{65.8} \\
		& StaticUI & &&& & && & &&& \\
	\hline
		\multirow{5}{*}{$1.0$} & Non-hier & \resulta{82.0}{0.84}{97.8}{92.9} &\resultb{83.7}{0.86}{97.7} &\resulta{83.2}{0.85}{97.4}{100.0 \\
		& Static & \resulta{40.2}{0.49}{82.5}{60.3} &\resultb{57.1}{0.74}{76.8} &\resulta{73.3}{1.08}{67.8}{72.3} \\
		& StaticUI & \resulta{73.3}{0.66}{110.4}{24.9} &\resultb{71.5}{0.86}{82.9} &\resulta{104.8}{1.53}{68.3}{88.6} \\
		& Dynamic & \resulta{64.0}{0.63}{102.1}{81.7} &\resultb{73.9}{0.75}{98.5} &\resulta{79.5}{0.82}{97.5}{99.7} \\
		& DynamicUI & \resulta{72.5}{0.73}{100.0}{83.5} &\resultb{79.2}{0.81}{97.4} &\resulta{81.6}{0.83}{98.7}{97.6} \\
	\cline{2-13}
		& Non-labled & \resultaa{46.1}{0.49}{95.0}{67.1}} &\resultbb{65.0}{0.74}{87.7} &\resultaa{79.9}{0.98}{81.6}{86.6} \\
		& StaticUI & &&& & && & &&& \\
		\bottomrule

	\end{tabular}
\end{adjustbox}
\end{table}

\newpage

\begin{table}[!ht]
\centering
	\caption{Results on the rare distracting test of: DAS ratio, average AS of distracting utterances (DAS) (\%), average AS of original utterances in {\em History} (Avg.) (\%), and AS of the first/last utterance in {\em History} (\%).}
	\label{table:rare}
\begin{adjustbox}{angle=90}
	\begin{tabular}{c|c|cccc|ccc|cccc}
	\toprule
	    \multicolumn{2}{c|}{Model} & \multicolumn{11}{c}{Distracting Test Set}\\
	\hline
		\multirow{2}{*}{Probability} & \multirow{2}{*}{Structure} & \multicolumn{4}{c|}{Rare: Begin} & \multicolumn{3}{c|}{Rare: Middle} & \multicolumn{4}{c}{Rare: End}\\
		& & \resulta{DAS}{DAS ratio}{Avg.}{1st} & \resultb{DAS}{DAS ratio}{Avg.} & \resulta{DAS}{DAS ratio}{Avg.}{{Last}}\\
	\hline
		\multirow{5}{*}{$0.0$} & Non-hier & \resulta{74.7}{0.80}{93.8}{82.3} & \resultb{84.3}{0.92}{92.1} & \resulta{92.0}{1.01}{91.2}{96.7} \\
		& Static & \resulta{29.3}{0.37}{79.6}{47.9} & \resultb{55.5}{0.77}{72.2} & \resulta{77.8}{1.21}{64.3}{75.2} \\
		& StaticUI & \resulta{22.9}{0.30}{76.4}{42.2} & \resultb{51.1}{0.75}{67.9} & \resulta{74.2}{1.22}{61.0}{73.2} \\
		& Dynamic & \resulta{61.4}{0.66}{93.6}{72.7} & \resultb{77.4}{0.89}{87.2} & \resulta{87.0}{1.06}{82.3}{85.4} \\
		& DynamicUI & \resulta{73.8}{0.73}{100.5}{81.0} & \resultb{87.9}{0.93}{94.2} & \resulta{90.1}{0.97}{93.0}{93.0} \\
	\hline
		\multirow{6}{*}{$0.5$} & Non-hier & \resulta{62.9}{0.69}{90.8}{84.0} & \resultb{72.6}{0.81}{89.2} & \resulta{77.7}{0.86}{90.3}{98.6} \\
		& Static & \resulta{29.4}{0.34}{86.1}{47.4} & \resultb{55.4}{0.71}{78.4} & \resulta{72.2}{0.99}{72.5}{82.6} \\
		& StaticUI & \resulta{24.7}{0.40}{62.2}{44.3} & \resultb{\cellcolor{r}\textbf{40.4}}{\cellcolor{r}\textbf{0.69}}{\cellcolor{r}\textbf{58.6}} & \resulta{53.0}{0.96}{55.4}{60.5} \\
		& Dynamic & \resulta{58.1}{0.61}{95.4}{76.7} & \resultb{70.3}{0.77}{91.0} & \resulta{76.0}{0.85}{89.4}{92.2} \\
		& DynamicUI & \resulta{60.9}{0.61}{100.5}{77.7} & \resultb{75.5}{0.80}{94.9} & \resulta{78.6}{0.83}{94.4}{93.6} \\
	\cline{2-13}
		& Non-labled & \resultaa{36.4}{0.40}{90.4}{53.7} & \resultbb{64.4}{0.80}{80.7} & \resultaa{82.7}{1.11}{74.6}{87.4} \\
		& StaticUI & &&& & && & &&& \\
	\hline
		\multirow{6}{*}{$0.7$} & Non-hier & \resulta{64.1}{0.71}{90.5}{83.1} & \resultb{75.0}{0.85}{88.8} & \resulta{78.1}{0.87}{89.6}{97.3} \\
		& Static & \resulta{30.0}{0.41}{73.5}{48.3} & \resultb{48.7}{0.70}{69.4} & \resulta{63.6}{0.98}{64.6}{71.7} \\
		& StaticUI & \resulta{20.9}{0.36}{57.7}{37.4} & \resultb{37.6}{0.70}{54.1} & \resulta{\cellcolor{r}\textbf{50.6}}{0.99}{\cellcolor{r}\textbf{51.1}}{\cellcolor{r}\textbf{57.3}} \\
		& Dynamic & \resulta{55.8}{0.58}{96.6}{73.9} & \resultb{67.8}{0.73}{92.4} & \resulta{74.6}{0.83}{89.8}{92.5} \\
		& DynamicUI & \resulta{59.7}{0.60}{98.8}{76.1} & \resultb{74.4}{0.80}{93.2} & \resulta{75.9}{\cellcolor{r}\textbf{0.81}}{93.9}{93.8} \\
	\cline{2-13}
		& Non-labled & \resultaa{30.8}{0.43}{70.9}{51.0} & \resultbb{48.1}{0.73}{66.3} & \resultaa{60.5}{0.97}{62.5}{67.3} \\
		& StaticUI & &&& & && & &&& \\
	\hline
		\multirow{6}{*}{$1.0$} & Non-hier & \resulta{82.8}{0.85}{97.8}{92.9} & \resultb{84.8}{0.87}{97.5} & \resulta{85.6}{0.88}{97.3}{100.2} \\
		& Static & \resulta{37.5}{0.46}{81.6}{60.3} & \resultb{55.0}{0.71}{77.4} & \resulta{65.2}{0.88}{74.4}{79.8} \\
		& StaticUI & \resulta{\cellcolor{r}\textbf{22.4}}{\cellcolor{r}\textbf{0.21}}{\cellcolor{r}\textbf{105.6}}{\cellcolor{r}\textbf{24.9}} & \resultb{71.0}{0.86}{83.0} & \resulta{103.1}{1.50}{68.7}{89.0} \\
		& Dynamic & \resulta{65.6}{0.65}{101.4}{81.7} & \resultb{75.1}{0.77}{98.1} & \resulta{79.6}{0.82}{96.9}{98.5} \\
		& DynamicUI & \resulta{74.3}{0.75}{99.3}{83.5} & \resultb{83.9}{0.88}{95.4} & \resulta{84.2}{0.88}{96.0}{94.3} \\
	\cline{2-13}
		& Non-labled & \resultaa{45.5}{0.49}{93.7}{67.1} & \resultbb{67.4}{0.77}{87.3} & \resultaa{80.6}{0.98}{82.5}{87.4} \\
		& StaticUI & &&& & && & &&& \\
    \bottomrule
	\end{tabular}
\end{adjustbox}
\end{table}

\newpage

\subsection{Detailed Results on the Distracting Tests}

In addition to DAS ratio, Table~\ref{table:random} shows the average AS of distracting utterances and of original utterances in {\em History}. Table~\ref{table:frequent} and Table~\ref{table:rare} additionally show the AS of the first or last utterances in {\em History}. Note again that an attention score of $100\%$ for a utterance indicates that this utterance receives an average attention score, e.g. for a dialogue containing 10 utterances, an attention score of $100\%$ indicates that the utterance receives $10\%$ attention out of all.  

From Table~\ref{table:random} it is clear that the average AS of the original utterances in {\em History} varies by model variants. A higher average AS for {\em History} indicates a lower AS for {\em Query}. Some models distribute most of the attention to {\em Query} while some models distribute the attention evenly to both {\em History} and {Query}. Normally, {\em Query} contains more relevant information, so we expect a lower average AS for {\em History}; however, the average AS for {\em History} is not the lower the better, since there are still some utterances in {\em History} that are important for the context. A lower average AS for {\em History} comes together with a lower average AS for distracting utterances (or a lower DAS), so DAS ratio is better suited for evaluating a model's capability on context attention distribution, since it takes the average AS for original utterances in {\em History} into account. In Table~\ref{table:random}, the models with the lowest DAS ratio also have the lowest average AS for distracting utterances and original utterances, while in Table~\ref{table:frequent} and Table~\ref{table:rare}, it is not always the case.

In Table~\ref{table:frequent} and Table~\ref{table:rare}, for the distracting test sets where distracting utterances are put in the beginning/end of the context, we show AS for the first/last utterance in {\em History} to have a clearer comparison. We can see in columns of Frequent: Begin and Rare: Begin that the distracting utterances usually receive lower attention than the first utterance in {\em History}, while the other original utterances in {\em History} receive more attention than the first utterance. This indicates a good performance of the model variants. Utterances far away from {\em Query} are normally distributed lower attention, so in a normal case, it is natural that the utterances that come after the first utterance receive more attention; however, these distracting utterances receive less attention, regardless of the fact that they are placed after the first utterances. It can thus be inferred that most model variants can distinguish distracting utterances as unimportant and distribute less attention to them. Similarly, the last utterances in {\em History} usually get more attention, while as the columns of Frequent: End and Rare: End show, distracting utterances receive less attention compared to other original utterances in {\em History}, regardless of that the distracting utterances are placed closer to {\em Query}.

\subsection{Summary of Results}

DAS ratio can distinguish conversational agents with similar perplexity on their ability of context attention distribution. In general, models trained with our proposed optimization strategy focus less on distracting utterances and more on original utterances in {\em History}. For most models, DAS ratios decrease by about 10$\%$ when trained with our proposed strategy with a $0.5$ or $0.7$ probability level. $0.7$ is generally the best option for a training inserting probability.

% Non-atten-loss models perform better than baselines while usually worse than those trained with attention loss.

% Rare distractions are as hard as frequent distractions for models to distinguish, especially when they are located near {\em Query}.

% Training inserting probabilities of $0.7$ and $0.5$ are better options for hierarchical models than $1.0$. The Non-hier models have better performance on $0.5$, while the StaticUI models have better performance on $0.7$. Probability $1.0$ is usually not a good choice. With probability $1.0$, models often assign either very low or very high attention to the distracting utterances, which results in extremely good or bad performance. Non-atten-loss models perform worse than those trained with attention loss, while they are better than baselines.

\section{Conclusions and Future Works}

We have studied context attention distribution, an essential component of multi-turn modelling for open-domain conversational agents. We have proposed an evaluation metric for context attention distribution based on the distracting test: DAS ratio. We have also improved the performance of context attention distribution for common multi-turn conversational agents through an optimization strategy via reducing the attention loss of self-contained distracting utterances. Extensive experiments show that our proposed strategy achieves improvements on most models, especially with a training inserting probability level of $0.7$. Future works can focus on adapting the proposed evaluation metric and optimization strategy to transformer-based conversational agents.

\newpage

{\Large\bfseries\scshape{Acknowledgements}}

This paper is funded by the collaborative project of DNB ASA and Norwegian University of Science and Technology (NTNU). We also received assist on computing resources from the IDUN cluster of NTNU \cite{IDUN}. We would like to thank Benjamin Kille and Peng Liu for their helpful comments.

\bibliographystyle{IEEEtran}
\bibliography{references}

\end{document}